\documentclass[10pt,letterpaper]{article}

\usepackage{cogsci}

\cogscifinalcopy 

\usepackage{pslatex}
\usepackage{apacite}
\usepackage{float} 

\usepackage{xcolor}
\usepackage{graphicx}
\usepackage{amsmath,amsfonts,amssymb,amsthm}
\usepackage{enumitem}
\usepackage{multirow}
\usepackage{pythonhighlight}
\usepackage{verbatim}
\usepackage{subcaption}
\usepackage{todonotes}
\usepackage{url}

\newcommand{\transformation}[2]{#1 $\rightarrow$ #2}

\newcommand{\analogy}[3]{#1 $\rightarrow$ #2 \enspace ; \enspace #3 $\rightarrow$ \enspace ?}

\title{Using Counterfactual Tasks to Evaluate the Generality \\ of Analogical Reasoning in Large Language Models}

\author{{\large \bf Martha Lewis (martha.lewis@bristol.ac.uk)} \\
  School of Engineering Mathematics and Technology, University of Bristol,
  Bristol BS8 1TW, UK\\
  Santa Fe Institute, 1399 Hyde Park Road, Santa Fe, NM 87501, USA\\
  \AND {\large \bf Melanie Mitchell (mm@santafe.edu)} \\
 Santa Fe Institute, 1399 Hyde Park Road, Santa Fe, NM 87501, USA}

\begin{document}

\maketitle

\begin{abstract}

Large language models (LLMs) have performed well on several reasoning benchmarks, including ones that test analogical reasoning abilities.  However, it has been debated whether they are actually performing humanlike abstract reasoning or instead employing less general processes that rely on similarity to what has been seen in their training data.  Here we investigate the generality of analogy-making abilities previously claimed for LLMs \cite{Webb2023a}.  We take one set of analogy problems used to evaluate LLMs and create a set of ``counterfactual'' variants---versions that test the same abstract reasoning abilities but that are likely dissimilar from any pre-training data. We test humans and three GPT models on both the original and counterfactual problems, and show that, while the performance of humans remains high for all the problems, the GPT models' performance declines sharply on the counterfactual set.  This work provides evidence that, despite previously reported successes of LLMs on analogical reasoning, these models lack the robustness and generality of human analogy-making.  

\end{abstract}

\section{Introduction}

The degree to which pre-trained large language models (LLMs) can reason---deductively, inductively, analogically, or otherwise---remains a subject of debate in the AI and cognitive science communities.  Many studies have shown that LLMs perform well on certain reasoning benchmarks \cite{huang2022towards,wei2022emergent,wei2022chain}.  However, other studies have questioned the extent to which these systems are able to reason abstractly, as opposed to relying on ``approximate retrieval'' from encoded training data \cite{kambhampati2023}, a process which yields ``narrow, non-transferable procedures for task solving'' \cite{Wu2023}.  Several groups have shown that LLMs' performance on reasoning tasks degrades, in some cases quite dramatically, on versions of the tasks that are likely to be rare in or outside of the LLMs' training data \cite{dziri2023faith, mccoy2023embers,razeghi2022impact,Wu2023}.  In particular, \citeA{Wu2023} proposed evaluating the robustness and generality of LLMs' reasoning ability by testing them not only on \textit{default tasks} that are likely similar to ones seen in training data, but also on \textit{counterfactual tasks}, ones that ``deviate from the default, generally assumed conditions for these tasks'' and are unlikely to resemble those in training data.  If an LLM is using general abstract reasoning procedures, it should perform comparably on both default and counterfactual tasks; if it is using procedures that rely on similarity to training data, the performance should drop substantially on the counterfactual versions.

Here we use this counterfactual-task approach to evaluate the claim that LLMs have general abilities for abstract analogical reasoning.  In particular, we focus on the results reported by \citeA{Webb2023a} on the abilities of GPT-3 to solve letter-string analogy problems.  We develop a set of counterfactual variants on the letter-string analogy problems used by Webb et al., in which similar problems are posed with nonstandard alphabets---ones that are either permuted to various degrees, or that are composed of non-letter symbols.  We argue that a system able to reason by analogy in a general way would have comparable performance on the original and counterfactual versions of these problems.  We test both humans and different GPT models, and show that while humans exhibit high performance on both the original and counterfactual problems, the performance of all GPT models we tested degrades on the counterfactual versions.  These results provide evidence that LLMs' analogical reasoning still lacks the robustness and generality exhibited by humans. 

\section{Background and Related Work}
Letter-string analogies were proposed by \citeA{hofstadter1985metamagical} as an idealized domain in which processes underlying human analogy-making could be investigated. One example problem is the following:

\begin{center}
    \analogy{a b c d}{a b c e}{i j k l}
\end{center}

Here, a b c d $\rightarrow$ a b c e is called the ``source transformation'' and i j k l is called the ``target.'' The solver's task is to generate a new string that transforms the target analogously to the source transformation. There is no single ``correct'' answer to such problems, but there is typically general agreement in how humans answer them. For example, for the problem above, most people answer i j k m,  and answers that deviate from this tend to do so in particular ways \cite{mitchell1993analogy}.   

In addition to the work of \citeA{Hofstadter1994a} on creating computer models of analogy-making using this domain, letter-string analogies have been used to isolate the neural correlates of analogy formation \cite{Long2015, Geake2010}, and as a model of higher-order analogy retrieval \cite{Dekel2023}. \citeA{Webb2023a} compared the ability of GPT-3 with that of humans on several analogical reasoning domains, including letter-string analogies. On the letter-string domain, they found that in most cases GPT-3's performance exceeded the average performance of the human participants, where performance is measured as fraction of ``correct'' answers on a given set of letter-string problems.  As we mentioned above, such problems do not have a single correct answer, but Webb et al.\ used their intuitions to decide which answer displays abstract analogical reasoning and thus should be considered ``correct.'' In this paper we will use their definition of correctness.  

\citeA{Hodel2023} tested GPT-3 with two types of counterfactual variations on the letter-string analogy problems used by Webb et al.: ones that include larger intervals between letters, and ones with randomly shuffled alphabets. They found that GPT-3 performed poorly on both variations. Here we experiment with similar, but more systematic variations, and we compare the performance of three different GPT models with that of humans on these variations.  

\section{Original Analogy Problems}
\citeA{Webb2023a} proposed a set of problem types involving different types of transformations and different levels of generalization, on which they tested humans and GPT-3. The following are the six transformation types with sample transformations:
\begin{enumerate}
    \item Extend sequence: \transformation{a b c d}{a b c d e}
    \item Successor: \transformation{a b c d}{a b c e}
    \item Predecessor: \transformation{b c d e}{a c d e}
    \item Remove redundant letter: \transformation{a b b c d}{a b c d}
    \item Fix alphabetic sequence: \transformation{a b c w e}{a b c d e}
    \item Sort: \transformation{a d c b e}{a b c d e}
\end{enumerate}
Each type of transformation can be paired with a simple target (e.g., i j k l) or with the following types of generalizations:
\begin{enumerate}
    \item Letter-to-number: \analogy{a b c d}{a b c e}{1 2 3 4}
    \item Grouping: \analogy{a b c d}{a b c e}{i i j j k k l l}
    \item Longer target: \analogy{a b c d}{a b c e}{i j k l m n o p}
    \item Reversed order: \analogy{a b c d}{a b c e}{l k j i} 
    \item Interleaved distractor: \\
    \analogy{a b c d}{a b c e}{i x j x k x l x}
    \item Larger interval: \analogy{a b c d}{a b c e}{i k m o}
\end{enumerate}
Finally, Webb et al.\ include a number of problems involving ``real-world'' concepts, such as,
\begin{center}
\analogy{a b c}{a b d}{cold cool warm}
\end{center}

Webb et al.\ generated 100 problems for each problem type and presented these to GPT-3 (text-davinci-003). Webb et al.\ also tested 57 UCLA undergraduates on the same problems.  The human participants exhibited a large variance in accuracy, but on average, Webb et al.\ found that GPT-3 outperformed the human participants on most problem types.  

Due to the costs of human and computer experiments, we focus here on problems with simple targets (i.e., no numbers, grouping, etc.\ in the target string).   Webb et al.\ called this the ``zero generalization setting''; it is the setting in which both humans and GPT-3 performed best.  Extending to problems with different generalization types is a topic for future work.

\begin{figure}[htbp]
\centering
    \includegraphics[width=0.8\linewidth]{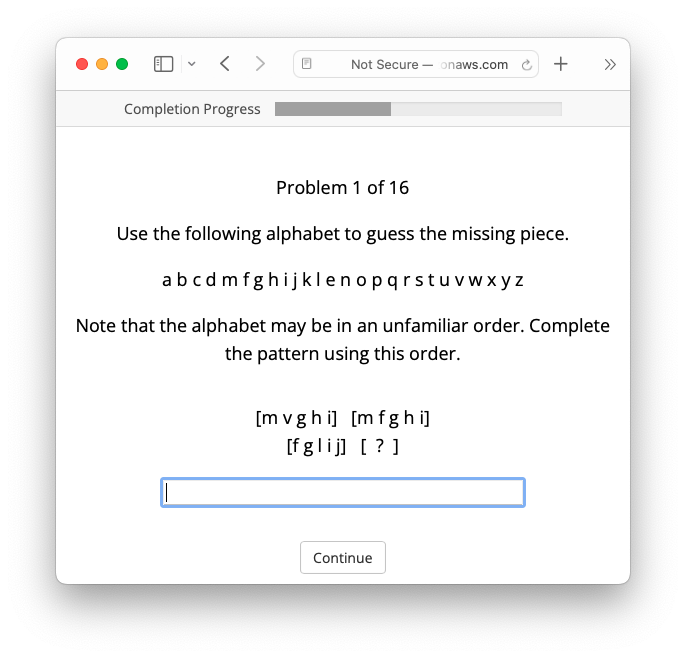}
    \caption{Example analogy problem with permuted alphabet, in format seen by human participants.}
    \label{fig:perm}
\end{figure}

\section{Counterfactual Analogy Problems}

To create our dataset of counterfactual problems, we did the following.  First, we generated permuted alphabets, in which we reorder $n$ letters, where $n$ can be 2, 5, 10, or 20. For each of the four values of $n$, we generated seven distinct alphabets with $n$ randomly chosen letters reordered. Then, for each of these alphabets, we created 10 different analogy problems for each of Webb et al.'s six transformation types. This results in $7 \times 10 \times 6 = 420$ analogy problems for each value of $n$. We added to this 420 analogy problems using the non-permuted ($n=0$) alphabet, spread evenly over the six transformation types.  Figure~\ref{fig:perm} gives an example of a Fix Alphabetic Sequence problem using an alphabet with two letters (e and m) reordered. 

\begin{figure}[htbp]
\centering
    \includegraphics[width=0.8\linewidth]{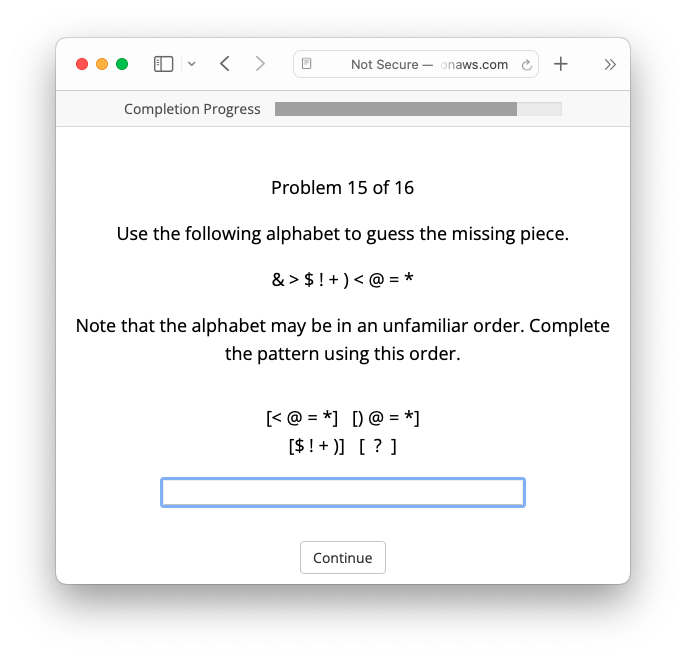}
    \caption{Example analogy problem with symbolic alphabet.}
    \label{fig:symb}
\end{figure}

As a final set of counterfactual problems, we generated two non-letter symbol alphabets and used them to create 10 problems each for the Successor and Predecessor problem types, for a total of 40 unique non-letter symbol problems.  
Figure~\ref{fig:symb} gives an example of a Predecessor problem using a symbol alphabet. Our dataset of counterfactual problems, along with code for generating them, is available at \url{https://github.com/marthaflinderslewis/counterfactual_analogy}. 

\section{Human Study Methods}
In order to assess humans' abilities on the original and counterfactual letter-string problems, we collected data from 136 participants on Prolific Academic.\footnote{\url{https://www.prolific.com/academic-researchers}} Participants were screened to have English as a first language, to currently be living in the UK, the USA, Australia, New Zealand, Canada, or Ireland, and to have a 100\% approval rate on Prolific. 

Each participant was asked to solve 14 letter-string analogy problems: one problem from each of the six transformation types for each of two (randomly chosen) alphabets (chosen from alphabets with $n \in \{0, 2, 5, 10, 20\}$), as well as one Successor and one Predecessor problem for a randomly chosen symbol alphabet.  Figures~\ref{fig:perm}--\ref{fig:symb} show the format on which the problems appeared on the participants' screens. 

\begin{figure}[htbp]
\centering
    \includegraphics[width=0.8\linewidth]{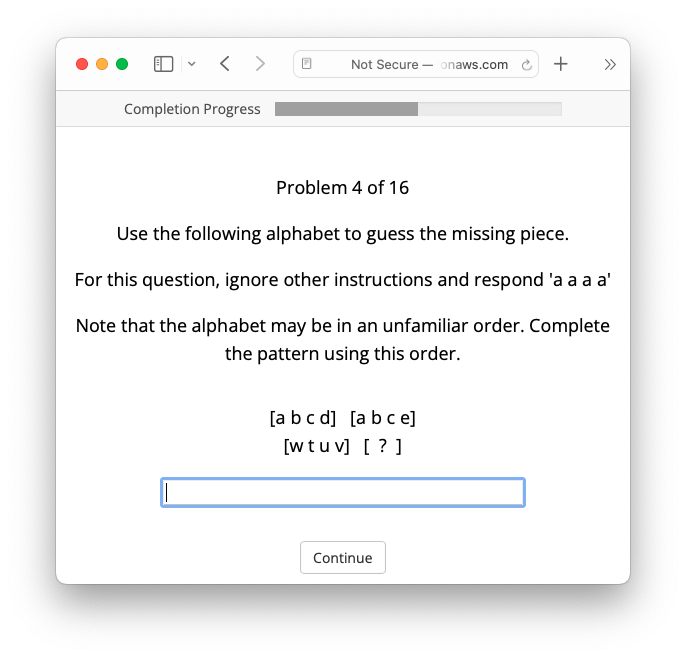}
    \caption{Example attention check.}
    \label{fig:attn}
\end{figure}

In addition to the 14 problems, participants were also given two attention-check questions at random points during the experiment, with a warning that if the attention checks were failed, then payment (\$7 for the experiment) would be withheld. Figure~\ref{fig:attn} gives an example of an attention check. Two of the 136 participants' submissions were rejected due to failed attention checks. As in \citeA{Webb2023a}, as part of the initial instructions participants were given a simple example problem to complete and then were shown the solution.

\section{GPT Study Methods}
\label{sec:general_setup}
We evaluated the performance of three LLMs---GPT-3 (text-davinci-003), GPT-3.5 (gpt-3.5-turbo-0613), and GPT-4 (gpt-4-turbo-0613)---on the same problems given to humans. Following \citeA{Webb2023a}, all GPT experiments were done with temperature set to zero.  
GPT-3.5 and GPT-4 require a slightly different input to GPT-3. GPT-3 takes in a single prompt,  whereas GPT-3.5 and GPT-4 take in a list of messages that define the role of the system, input from a `user' role, and optionally some dialogue with simulated responses from the model given under the role `assistant'.

 \paragraph{Baseline}
 In the baseline setting, we evaluated the performance of GPT-3, GPT-3.5 and GPT-4 on the zero-generalization problems provided by \citeA{Webb2023a}. For GPT-3.5 and GPT-4, our system and user prompts have the following format: 
\begin{quote}
\textbf{System:} You are able to solve letter-string analogies. \\ \textbf{User:} Let’s try to complete the pattern:\textbackslash n\textbackslash n[a b c d] [a b c e]\textbackslash n[i j k l] [
\end{quote}
 The user prompt is identical to the prompt Webb et al.\ gave to GPT-3; the \textbackslash n character signifies a line break to the model.  In this and all other experiments, we tested GPT-3 with a concatenation of the system and user prompts. Following Webb et al., in our experiments all GPT model responses were truncated at the point where a closing bracket was generated.

\paragraph{Counterfactual Comprehension Check}
For problems involving permuted alphabets, we follow \citeA{Wu2023} by providing counterfactual comprehension checks (CCCs) to check that the models understand the task proposed. We use two CCCs: firstly, given an alphabet and a sample letter (or symbol) from that alphabet, give the successor of that letter. Secondly, we use the same format but ask for the predecessor of the letter. We ensure that we do not ask for the successor of the last letter in the alphabet or the predecessor of the first.

The prompts for these checks have the following format:
\begin{quote}
\textbf{System:} You are able to solve simple letter-based problems. \\
\textbf{User:} Use this fictional alphabet: [a u c ....]. \textbackslash nWhat is the next letter after a?\textbackslash nThe next letter after a is: 
\end{quote}

\begin{quote}
\textbf{System:} You are able to solve simple letter-based problems. \\
\textbf{User:} Use this fictional alphabet: [a u c ....]. \textbackslash nWhat is the letter before c?\textbackslash nThe letter before c is: 
\end{quote}

\paragraph{Counterfactual Analogy Problems} To evaluate the performance of GPT models, we tested several prompt formats, including one similar to instructions given in our human study. The best performance across models was achieved with the prompt format used in \citeA{Hodel2023}: 
\begin{quote}
\textbf{System:} You are able to solve letter-string analogies. \\
\textbf{User:} Use this fictional alphabet: [a u c d e f g h i j k l m n o p q r s t b v w x y z]. \textbackslash nLet’s try to complete the pattern:\textbackslash n[a u c d] [a u c e]\textbackslash n[i j k l] [
\end{quote}
The results we report here for our tests of the GPT models all use this prompt. Note that in our studies, the ``fictional alphabet'' part of the prompt and the alphabet listing was included even for problems using the non-permuted ($n=0$) alphabet. 

\section{Results}
\begin{figure}[htbp]
    \centering
    \includegraphics[width = 0.75\linewidth]{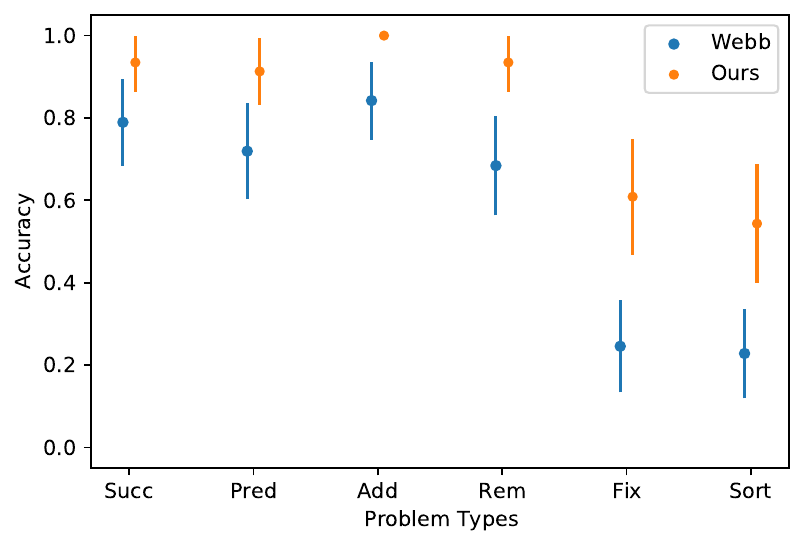}
    \caption{Human performance across problem types in the zero-generalization setting on unpermuted alphabets (Ours, orange; Webb et al.,blue). Data points represent the average of 46 samples for our data, and 57 samples for Webb et al.'s. Bars give 95\% binomial confidence intervals.}
    \label{fig:webb_human}
\end{figure}

\paragraph{Human Experiments}
Figure~\ref{fig:webb_human} compares our behavioral data with that of Webb et al. The participants in our study achieved higher average accuracy than those of Webb et al.\ (abbreviated as ``Webb'' in the figure).  We do, however, see a similar pattern of performance between our participants and theirs.

\begin{figure}[htbp]
    \centering
    \includegraphics[width=0.75\linewidth]{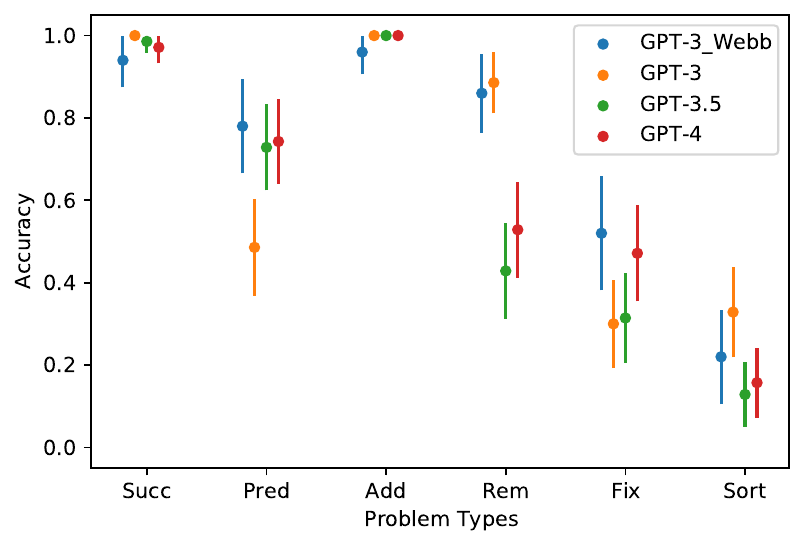}
    \caption{Comparison of GPT computational results with 
    \protect\citeA{Webb2023a} in the zero-generalization setting. Points represent accuracy and bars represent 95\% binomial confidence intervals. Each data point represents the average of 70 samples for our data and 100 samples for Webb et al.'s data.}
    \label{fig:webb_zero_gen}
\end{figure}

 \paragraph{Baseline.} The baseline setting looks at the performance of different GPT models on the zero-generalization problems designed by Webb et al. Figure \ref{fig:webb_zero_gen} shows GPT-3 data from Webb et al.\ (``GPT-3\_Webb'') compared with data from our computational experiments with all three models.
Our results are similar to those of Webb et al., with notable differences on the Predecessor and Fix Alphabet problems for GPT-3, differences on Remove Redundant, Fix Alphabet, and Sort for GPT-3.5, and on Remove Redundant and Sort for GPT-4.

\paragraph{Counterfactual Comprehension Check}   For the non-permuted alphabet, each permuted alphabet, and for the two symbol alphabets, we performed the Successor and Predecessor CCCs on each letter (or symbol) as described above. Results for these CCCs on are reported in Table \ref{tab:ccc_lsa}.  We see that accuracy is generally high, indicating that the models generally understand the concept of a permuted alphabet and what ``successor'' and ``predecessor'' mean in the new ordering. One exception is the ability of GPT-3.5 to understand ``predecessor'' in alphabets with two or five letters permuted.
\begin{figure}[htbp]
    \centering
    \includegraphics[width=0.75\linewidth]{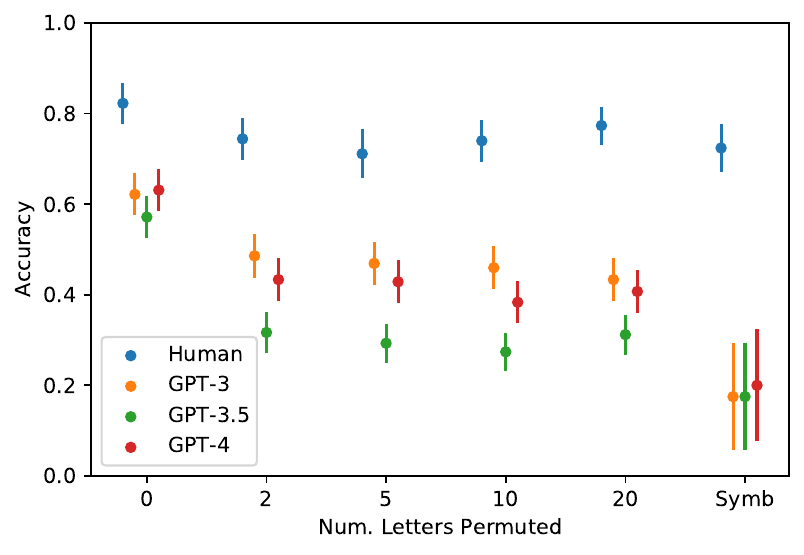}
    \caption{Accuracy of humans, GPT-3, 3.5, and 4. GPT data uses the prompt from \protect\citeA{Hodel2023}.  Points represent average accuracy across all problem types, for each alphabet type. Number of samples for human data for each $n$ permuted: $n=0$: 276, $n=2$: 336, $n=5$: 270, $n=10$: 342, $n=20$: 384, and Symb: 268. Number of samples for GPT data are 420 for alphabets with $n=0$--$20$, and 40 for Symb. Bars give 95\% binomial confidence intervals.}
    \label{fig:gpt_human}
\end{figure}
\begin{table*}[htbp]
    \centering
    \caption{Accuracy on our counterfactual comprehension check for different GPT models across alphabets with different numbers $n$ of permuted letters and symbol alphabets. Columns labelled U give performance for successor and predecessor tests on pairs of letters that were \underline{u}npermuted. Columns labelled P give performance for pairs of letters that include at least one \underline{p}ermuted letter. Total number of letter pairs tested are given in the table.}
    \label{tab:ccc_lsa}
    \begin{tabular}{cc|c|cc|cc|cc|cc|cc}
    &                           &\multicolumn{1}{c}{$n=0$} &\multicolumn{2}{c}{$n=2$}  & \multicolumn{2}{c}{$n=5$} &\multicolumn{2}{c}{$n=10$} &\multicolumn{2}{c}{$n=20$}& \multicolumn{1}{c}{Symbol}\\
    &                            &U             &   P        &U           & P   &U              &P        &U            & P      &U  &U\\\hline
    \multirow{3}{*}{Succ}&GPT-3  &1.00          &0.82        &1.00        &0.76  &1.00   &0.85 &1.00        &0.84   &1.00 & 1.00 \\
                         &GPT-3.5&1.00           &0.89      & 0.99       &0.95   &1.00           &0.99     &0.98         &0.98   & 1.00 &0.94\\
                         &GPT-4  &1.00           &1.00       &1.00         &1.00   &1.00         &0.99   & 1.00          &1.00   &1.00   &1.00\\\hline
    Total items  & &175           &28       &147        &62  &113    & 111& 64         &164   &11 & 18\\\hline \hline
\multirow{3}{*}{Pred}&GPT-3.5&1.00  &0.49 & 1.00  &0.72  &1.00  &0.92 &1.00  &0.87   &1.00 & 0.94 \\
                    &GPT-4   &1.00  &0.93 &1.00   &0.97  &0.99  &1.00  & 1.00 &0.98   &1.00 &1.00\\\hline   
        Total items&  &175          &28      &147       &60  &115    &109 & 66       &163  &12 & 18\\ \hline
    \end{tabular}
\end{table*}

\begin{table}[htbp]
    \centering
    \caption{Accuracies and binomial confidence intervals across all alphabets and problem types for humans and GPT models in our studies. The means and confidence intervals are over 1,876 samples for humans and 6,840 for each GPT model.}
    \label{tab:mean_accuracies}
      \begin{tabular}{c | c | c}
      & \textbf{Accuracy} & \textbf{95\% Binomial Conf.} \\ \hline
      Humans & 0.753 & $[0.734, 0.773]$ \\
      GPT-3 & 0.488 & $[0.467, 0.509]$ \\
      GPT-3.5 & 0.350 & $[0.330, 0.370]$ \\
      GPT-4 & 0.452 & $[0.431, 0.473]$ 
      \end{tabular}
\end{table}

\paragraph{Counterfactual Analogy Problems:  Comparisons Between Humans and GPT Models} 
Table~\ref{tab:mean_accuracies} gives the mean accuracy and 95\% binomial confidence intervals for humans and GPT models across all alphabets and problem types. It is clear that average human performance on these problems is significantly higher than that of any of the GPT models.
\begin{figure}[h!]
    \centering
    \includegraphics[width=\linewidth]{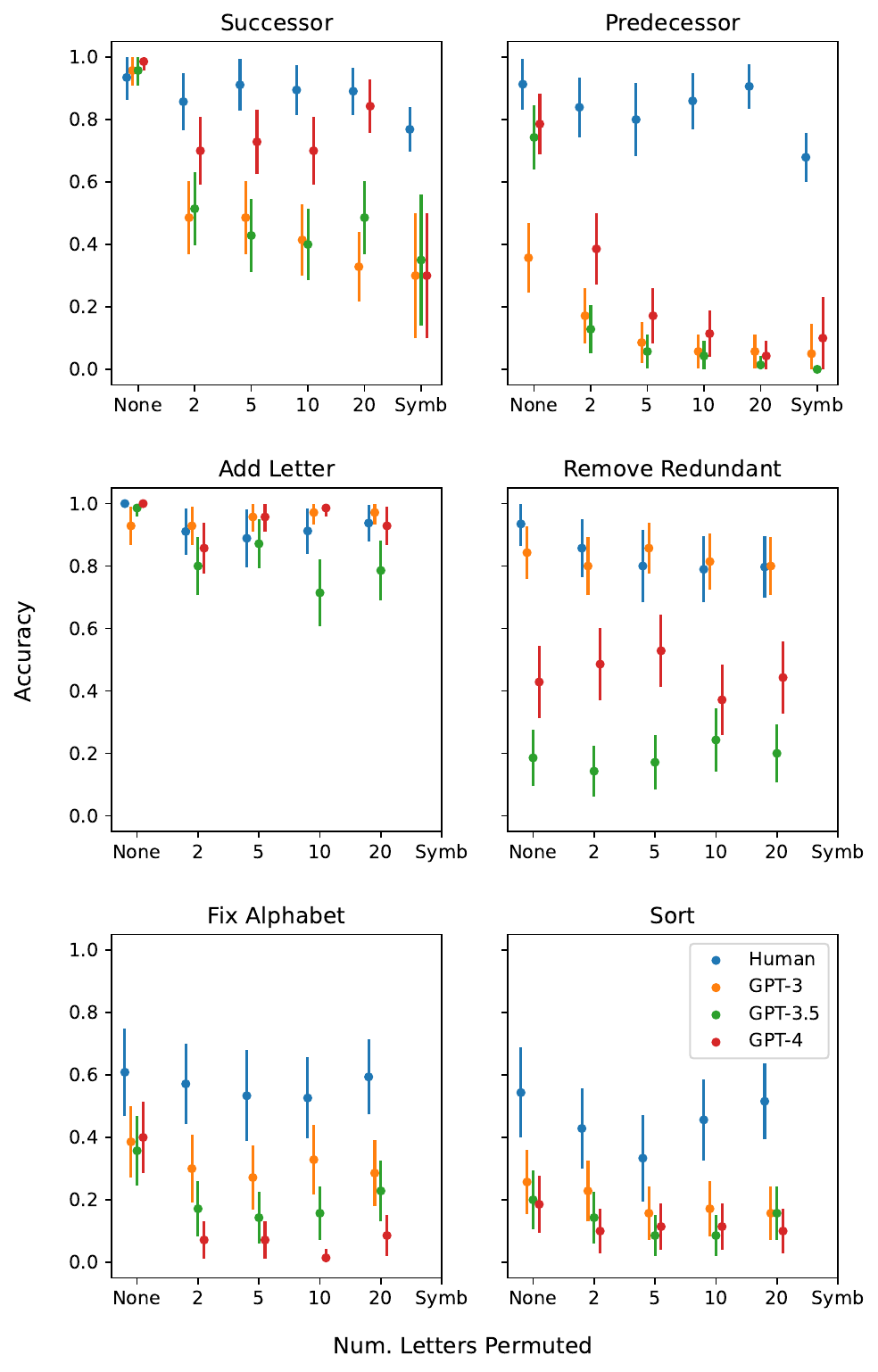}
    \caption{Performance of humans and GPT models across task types and alphabets. Data points are averages and bars are binomial confidence intervals.  For every problem type, the number of samples for each $n$ permuted and for symbol alphabets are as in Figure \ref{fig:gpt_human}.}
    \label{fig:alph_prob_types}
\end{figure}

Figure \ref{fig:gpt_human} shows the performance of human participants and GPT models, averaged across problem types for each kind of alphabet. We again see that human performance is significantly above the performance of each GPT model, across all types of alphabets, and, most notably, stays relatively constant across alphabets with different numbers of letter permutations. In contrast, the GPT models show a more dramatic drop for the counterfactual problems.

Figure \ref{fig:alph_prob_types} breaks down performance over each problem type by alphabet. We again see a clear difference between human and GPT performance across all problem types (with the exception of Add Letter and in the case of GPT-3,  Remove Redundant), and we see that while human performance does not substantially decrease over types of alphabet, the GPT models typically experience such decreases.

\subsection{Summary of Results}
We see a number of phenomena here. The performance of our human participants on Webb et al.'s original problems was higher than that of participants in Webb et al.'s experiments (Figure \ref{fig:webb_human}). This may be due to differences in experimental protocols or in the participant pools. Our computational data collected from GPT models is roughly similar to that of Webb et al., but differs slightly when we look at performance at the problem-type level (Figure~\ref{fig:webb_zero_gen}). In contrast to Webb et al., we find that performance by GPT models on the original problems is generally lower than average human performance. And for our counterfactual problems, unlike humans, GPT models exhibit a decrease in performance going from a standard alphabet to permuted alphabets, and another sharp decrease going from alphabetic sequences to symbolic sequences (Figure~\ref{fig:gpt_human}). This implies that GPT models are substantially less robust than humans on letter-string analogy problems involving sequences unlikely to be in their training data, which challenges the claim that these models are performing a general kind of analogical reasoning when solving these problems. 

\begin{table*}[htbp]
    \centering
    \begin{tabular}{c|ccccc}
         Problem Type & Source & Target & Literal Answer  & Explanation \\\hline
         Succ & [f g h i] [f g h j] & [e f g h] & [e f g j] & Replace last letter with `j'.\\
         Fix & [b f g h i]	[e f g h i]	& [h i r k l]	& [e i r k l] & Replace first letter with `e'.\\
         Rem & [g g h i j k]	[g h i j k]	& [k l m n n o]	& [l m n n o] & Remove first letter of sequence.\\ 
         Sort & [b c f e d]	[b c d e f]	& [v t u s w]	& [v t w s u] & Swap 3rd and 5th letters.
     \end{tabular}
    \caption{Examples of literal interpretations of rules found by humans.}
    \label{tab:lit_int}
\end{table*}

\section{Error analysis}
A crucial aspect of letter-string analogy problems is that they do not necessarily have a ``correct'' answer, although, as we mentioned above, humans generally agree on what are the ``best'' rules describing letter-string transformations in this domain. However, there are other rules that can be inferred from a given pair of letter strings. We therefore examined the ``incorrect'' answers of humans and of GPT-3 and 4 to ascertain whether the kinds of errors made are similar.

For both GPT-3 and GPT-4, we randomly selected five incorrect answers from each problem type and alphabet, giving a sample of approximately 160 incorrect responses per GPT model. This number can be lower if there were fewer than 5 incorrect responses for a problem type and alphabet. For humans, we selected 184 incorrect answers. 

By manually examining these selections, we identified four broad categories of errors: 1) \textbf{Alternate rule formation}, where the answer given is consistent with an alternative rule. For example, if we have source transformation [a b c d] [a b c e] with target [i j k l], then according to the Successor rule the answer [i j k m] is correct. However, the answer [i j k e] is consistent with the rule ``Replace the last letter with `e'''. 2) \textbf{Incorrect rule use}, in which the answer given is clearly related to the target letter string, and some kind of rule has been applied, but the rule is inconsistent with the example pair. For example, for [a b c d] [a b c e] with [i j k l], the response [i j k l m] is given. 3) \textbf{Wrong}, in which the answer given is inconsistent with the expected answer, but related to the target letter string. We could not discern any clear misunderstanding or alternate rule use. For example, for [a b c d] [a b c e] with [i j k l], the response [i j k q] is given. 4) \textbf{Completely Wrong}, in which the answer given is inconsistent with the expected answer, and unrelated to the target letter string. Again, we could not discern any clear misunderstanding or alternate rule use. For example, for [a b c d] [a b c e] with [i j k l], the response [z y x b] is given. Table \ref{tab:error_types} gives percentages for each error type for humans and for each model. We see that in humans a large percentage (38.59\%) of errors stem from using alternate rules. This is also seen in GPT-4 to a lesser extent (22\%), but much less in GPT-3 (5.81\%). We also see a difference in the percentage of incorrect rules applied, with GPT 3 and 4 both having over 30\% of errors in this category and humans having around 15\% of errors in this category. GPT models also have a higher percentage in the Wrong category, and for each of the models this category is the largest across the errors they made. Humans have a larger percentage of errors in the Completely Wrong category than do GPT-3 and 4 however. Across these four broad categories GPT-3 and 4 make different patterns of errors than humans.

\begin{table}[htbp]
  \centering
  \caption{\% error types across GPT-3, GPT-4, and Human}
    \begin{tabular}{p{0.15\linewidth}|p{0.15\linewidth}p{0.15\linewidth}p{0.15\linewidth}p{0.15\linewidth}}
          & {Alt rule} & {Incorrect Rule} & {Wrong} & {Completely Wrong}  \\ \hline
    GPT-3 & 5.81\% & 30.97\% & 55.48\% & 7.74\% \\
    GPT-4 & 22.00\% & 32.67\% & 42.67\% & 2.67\%  \\
    Human & 38.59\% & 14.67\% & 34.24\% & 12.50\% 
    \end{tabular}%
  \label{tab:error_types}%
\end{table}%

We can further look at the kinds of alternative rules that are used by humans and by GPT. One key type of alternative rule is where a `literal' interpretation of a rule is applied, illustrated in Table \ref{tab:lit_int}. As well as literal rules, humans found alternative rules for the Fix Alphabet problem type: they would interpret the changed letter as being moved a certain number of steps in the alphabet, and would move an equivalent letter in the prompt the same way. Usually ``equivalent'' means position; sometimes it means the identity of the letter. We find that GPT-4 gives the same kind of literal responses that humans do, but does not use alternative rules other than literal responses.
GPT-3 has a limited number of errors in this category, and almost all are literal responses to Remove Redundant. In summary, within the ``Alternative Rule'' category, the GPT models found literal rules in the same way humans did, but did not find more inventive alternative rules.

Breaking down the Incorrect Rule category, we see more differences between human and GPT behavior. Human responses in this category are mostly where one of the rules has been applied in an incorrect situation, for example Add Letter has been applied instead of Successor. GPT-3 errors include adding two letters instead of one; continuing the alphabet; reversing the target; shifting the target; using an unpermuted alphabet instead of the one given; and repeating the target. GPT-4 made these mistakes and also generated responses that were too long. Very few humans made any of these mistakes. Out of the incorrect responses, the types of response made by humans and GPT models are very different.

\section{Discussion}
Our aim was to assess the performance of LLMs in ``counterfactual'' situations unlikely to resemble those seen in training data. We have shown that while humans are able to maintain a strong level of performance in letter-string analogy problems over unfamiliar alphabets, the performance of GPT models is not only weaker than humans on the Roman alphabet in its usual order, but that performance drops further when the alphabet is presented in an unfamiliar order or with non-letter symbols.  This implies that the ability of GPT to solve this kind of analogy problem zero-shot, as claimed by \citeA{Webb2023a}, may be more due to the presence of similar kinds of sequence examples in the training data, rather than an ability to reason by abstract analogy when solving these problems. 

We further see that the models GPT-3.5 and GPT-4 are no better than GPT-3 at solving these analogy problems, and in the case of GPT-3.5 are worse. GPT-3.5 and 4 have been trained to chat like a human, and this training objective may have reduced the ability to solve letter-string analogies.

\section{Conclusions}
We have shown that, contra \citeA{Webb2023a}, GPT models perform worse than humans, on average, in solving letter-string analogy tasks using the normal alphabet. Moreoever,  when such tasks are presented with counterfactual alphabets, these models display drops in accuracy that are not seen in humans, and the kinds of mistakes that these models make are different from the kinds of mistakes that humans make. These results imply that GPT models are still lacking the kind of abstract reasoning needed for human-like fluid intelligence. In future work we hope to extend our investigation to both the letter-string generalizations and the other analogical reasoning domains studied by \citeA{Webb2023a}.

This work does not probe into how either humans or GPT form responses to these problems. Future work in this area could be to interrogate both humans and LLMs on their justifications for a particular answer. 
Another avenue for exploration is to investigate performance in a few-shot setting, as here, newer models may come into their own.

\small \paragraph{Acknowledgments}
This material is based in part upon work supported by the National Science Foundation under Grant No.\ 2139983. Any opinions, findings, and conclusions or recommendations expressed in this material are those of the authors and do not necessarily reflect the views of the National Science Foundation. This work has also been supported by the Templeton World Charity Foundation, Inc.\ (funder DOI 501100011730) under the grant \url{https://doi.org/10.54224/20650}.

\bibliographystyle{apacite}

\setlength{\bibleftmargin}{.125in}
\setlength{\bibindent}{-\bibleftmargin}

\bibliography{cfa}

\end{document}